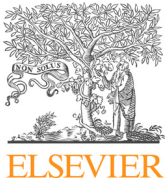

Contents lists available at ScienceDirect

# Neurocomputing

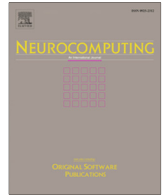

# Classification and generation of real-world data with an associative memory model

Rodrigo Simas [a,*], Luis Sa-Couto [a], Andreas Wichert [a]

[a] Instituto Superior Técnico, University of Lisbon, Portugal

A B S T R A C T

Drawing from memory the face of a friend you have not seen in years is a difficult task. However, if you happen to cross paths, you would easily recognize each other. The biological memory is equipped with an impressive compression algorithm that can store the essential, and then infer the details to match perception. The Willshaw Memory is a simple abstract model for cortical computations which implements mechanisms of biological memories. Using our recently proposed sparse coding prescription for visual patterns [34], this model can store and retrieve an impressive amount of real-world data in a fault-tolerant manner. In this paper, we extend the capabilities of the basic Associative Memory Model by using a Multiple-Modality framework. In this setting, the memory stores several modalities (e.g., visual, or textual) of each pattern simultaneously. After training, the memory can be used to infer missing modalities when just a subset is perceived. Using a simple encoder-memory-decoder architecture, and a newly proposed iterative retrieval algorithm for the Willshaw Model, we perform experiments on the MNIST dataset. By storing both the images and labels as modalities, a single Memory can be used not only to retrieve and complete patterns but also to classify and generate new ones. We further discuss how this model could be used for other learning tasks, thus serving as a biologically-inspired framework for learning.

© 2023 The Author(s). Published by Elsevier B.V. This is an open access article under the CC BY license (http://creativecommons.org/licenses/by/4.0/).

## 1. Introduction

Deep Artificial Neural networks have made remarkable progress in recent years, being able to solve extremely complex tasks at a human, or even "superhuman" level [17]. Despite their original inspiration in biological systems [30], these models no longer attempt to implement a general information processing system like that of the brain. Instead, these models serve as effective engineering tools that solve specific tasks with high accuracy. On the flip side, the field of Computational Neuroscience concerns itself with building more general models that are constrained by biological principles, not focused on specific tasks. Examples of these constraints include Hebb's postulate for the synaptic plasticity of cell assemblies [9], and the neural energy efficiency of sparse representations [23,24].

The basis for this work is the Willshaw model [37] of associative memory. This shallow artificial neural network is a likely candidate for a computational model of brain functions [9,27]: With extreme neural energy efficiency [15,18,20], this associate memory can efficiently store a tremendous amount of patterns [21,36,26,1,11]. The storage capacity of the basic model can even be further enhanced by implementing other neurological principles, such as structural plasticity [14]. However, this large capacity is only attainable when the stored patterns are sparse and randomly generated [13], which is not common in real-world data (see Fig. 2). In this paper, we apply the idea of multiple modalities [19,4] to the simple Willshaw Model. Without changing the basic architecture of the original model, our proposed framework allows the basic model to naturally solve more complex tasks, such as classification and gereration. Furthermore, our solution achieves the reported results with a very efficient Hebbian Training rule, which requires a single pass through the dataset to train the model for all the tasks.

Although this work is evaluated on the MNIST dataset for demonstration purposes, its purpose is not to outperform the highly-performing deep learning approaches. Instead, the goal of the Multiple-Modality architecture is not to solve a particular task, but to provide a flexible biologically-constrained framework such as the brain-like systems proposed by Hawkins [8]. All in all, the proposed architecture is not meant to be an accurate replica of the brain, but a simple framework that exhibits the flexibility to perform several complex tasks simultaneously, such as pattern completion, classification, and generation, under biological constraints.

---

* Corresponding author.
*E-mail addresses:* rodrigo.simas@tecnico.ulisboa.pt (R. Simas), luis.sa.couto@tecnico.ulisboa.pt (L. Sa-Couto), andreas.wichert@tecnico.ulisboa.pt (A. Wichert).

The rest of this paper is organized as follows. The remainder of Section 1 is dedicated to providing the necessary background on artificial associative memories and the Willshaw Model. On Section 2, we highlight the main advantages of the model and illustrate them with experiments on the MNIST dataset. Section 3 presents the new Multiple-Modality framework, and Section 4 analyses the performance of the new architecture as a classifier and generative model. Finally, on Section 5 we discuss the results and reflect on future possibilities enabled by this work.

### 1.1. Associative Memory

Associative Memories (AMs) are a family of Biologically-Inspired Artificial Intelligence models that imitate the mechanisms of biological memories [3]. These models learn by storing associations between pairs of patterns: Correlated features form synaptic connections between the memory's neurons. This way, a pattern is represented in the memory by the activation of a population of neurons. Each individual neuron can participate in many populations, thus allowing us to fill the memory to several factors of its number of units [28,32]. A trained memory can then be queried with a pattern: the neurons in the network will fire according to their learned connections, and the resulting population of active neurons will correspond to the memory's response to the query. This task is known as *Retrieval*, and the fact that the query is not an address, but a *content* vector with information, means that the memory is employing *content*-addressability.

### 1.2. Willshaw network

The Willshaw Network (WN) [37] is a shallow, feed-forward artificial neural network that stores a set of associations between question vectors $x$, and answer vectors $y$. The model implements a content-addressable AM that establishes a mapping $(x \rightarrow y)$.

The model is simply composed of $n$ neurons, each having $m$ binary connections to the input. Therefore, the model is fully represented by a binary matrix $W_{ij} \in \{0,1\} : i = 1, \ldots, m; j = 1, \ldots, n$, where the dimensions $m$ and $n$ are given by the fixed sizes of the *question* and *answer* vectors, respectively. This matrix defines the absence/presence of correlations between positions of the stored associations: A connection $W_{ij} = 1$ is formed during training when a correlation between the positions $i$ of a *question* vector, and the position $j$ of an *answer* vector is detected. Formally, given a set of $M$ pairs $(x, y)$, the weight matrix $W$ is computed as:

$$W_{ij} = \min\left(1, \sum_{\mu=1}^{M} x_i^\mu y_j^\mu\right). \tag{1}$$

Notice how the learning rule requires a single pass through the training set ($\sum_{\mu=1}^{M}$), and how a local Hebbian Rule [9] is employed ($x_i^\mu y_j^\mu$).

When a cue $\tilde{x}$ is shown as input to a trained memory, each neuron will fire according to its learned connections. Then, the memory will output its resulting state $\hat{y}$ which is denoted as the retrieved vector. The memory's response to the cue reflects the learned mapping $(x \rightarrow y)$, and since this mapping is implicitly stored through the connections of the network, the model is naturally able to generalize for novel or noisy cues. This process, denoted retrieval, can be formulated in two steps. First, the dendritic potential $s$, is computed for each neuron:

$$s_j = \sum_{i=1}^{m} W_{ij} \tilde{x}_i. \tag{2}$$

Then, the state of the network $\hat{y}$ is determined by applying the Heaviside step activation function $H$ to the potential of each neuron:

$$\hat{y}_j = H(s_j - \theta_j). \tag{3}$$

The threshold parameter $\theta_j$ will determine the sensitivity of the network. In this work we employ the *soft threshold* strategy [28,25] where $\theta_j = \max_{1 \leq i \leq n} s_i$.

A particularly interesting use case of the WN, is when we teach the model to learn the mapping $(x \rightarrow x)$, which is denoted *auto-association*. In such cases, the memory learns how to map a vector back to itself. In this setting, the memory has great practical uses such as reconstructing the whole pattern when only part of it is presented to the memory (Fig. 1).

The goal of an Associative Memory model is to store as much information as possible and retrieve it in a fault-tolerant manner. Despite the simplicity of the WN, it has been shown that this model, under specific conditions, is capable of storing a tremendous number of associations. More specifically, this is the case when the patterns that the model stores are Sparse Distributed Representations (SDRs) [2,26,27,29,10,6,24]. This type of representation, which closely resembles the selective neuron firing phenomenon of the human neocortex [7,24,22,11], corresponds to large binary vectors where only a subset of its bits are active (sparse), and where all of the positions of the vectors are quasi-uniformly used across a dataset (distributed). The main issue in the field of Associative Memories is the fact that real-world data is not naturally in the form of SDRs (Fig. 2).

Artificial Memories require sparse encoding prescriptions to transform natural data into SDRs. Again, an analogy with biological systems can be drawn. The brain does not process information directly in its raw format, instead, it relies on lower functional regions to transform sensory data into the sparse activation of neurons. For instance, the visual area of the brain is divided into several hierarchical regions: V1, V2, V4, and IT. The V1 area of the brain is responsible for detecting low-level visual features such as edges, and basic color [5]. The detection of such features results in the activation of a collection of neurons that forms a signal. This signal is passed onto the V2 area which will apply a similar process. After passing through all the regions, the final result will be an SDR that represents what is being perceived [8].

In our recent work [31,33], a sparse encoding prescription for visual patterns, coined the What-Where (WW) encoder, was proposed. This prescription is inspired by elements of the mammalian visual cortex [12] to produce informative binary compressions. Preliminary results on the MNIST dataset [16] have illustrated the quality of the WW codes in terms of their storage and error-tolerant retrieval [32]. In this paper, we go a step further and propose a biologically-constrained framework that allows a trained Willshaw Network to perform more complex learning tasks such as classification and generation.

## 2. Single Modality Willshaw Network

This section is dedicated to highlighting one of the main strengths of the Willshaw Network which is its ability to complete missing information. This feature of the basic architecture of the model is a key component of the Multiple-Modality architecture proposed in this paper, which will be presented in the upcoming section.

Our previous analysis [32] has focused on the retrieval process at the SDR level, where we compared the bits of the retrieval cues and retrieved vector to measure the memory's performance. The main conclusions of this study were:

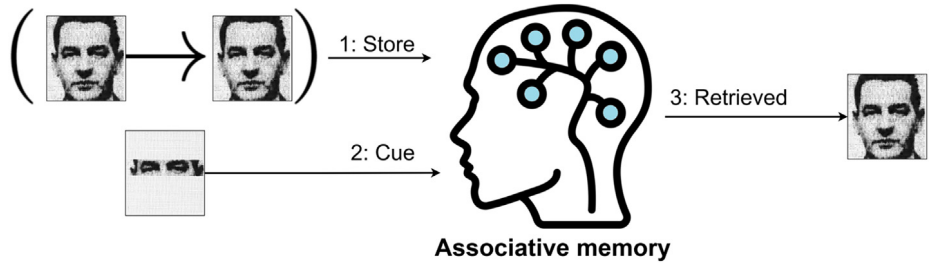

**Fig. 1.** Auto-association example. (1): An artificial Associative Memory model is used to store visual patterns of faces in auto-association ($x \rightarrow x$). (2): When an incomplete version of a stored pattern is shown to the memory as a retrieval cue ($\tilde{x}$), (3): the memory will retrieve ($\tilde{y}$) the original image.

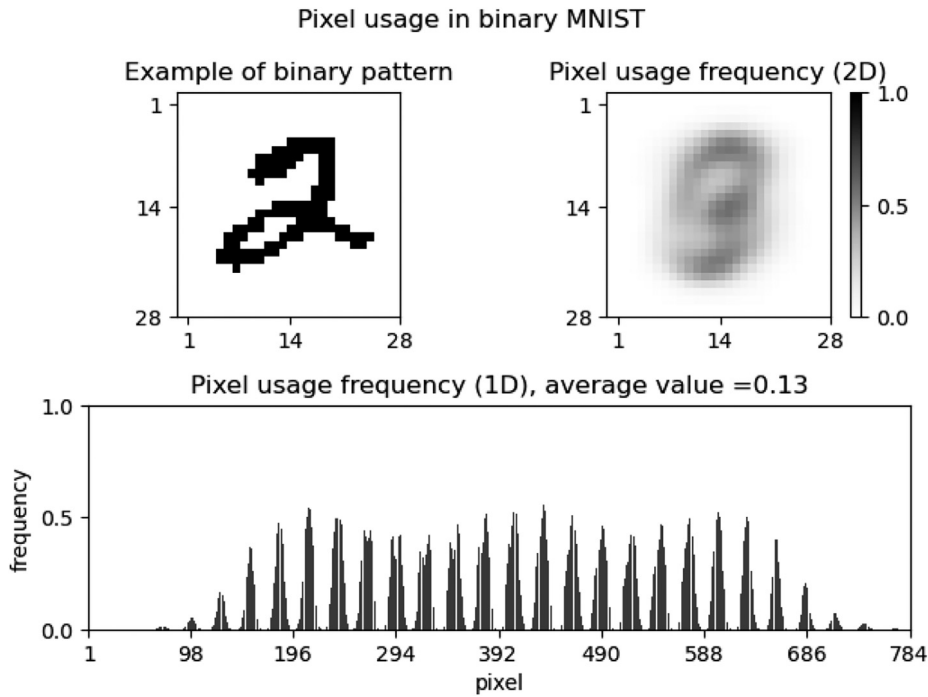

**Fig. 2.** The sparse coding problem. The Willshaw Network requires that the patterns it stores are sparse and distributed. For instance, [10] demonstrates that the optimal capacity is reached when the number of active bits $M$, in each pattern of size $n$, is $M = \log_2\left(\frac{n}{4}\right)$, and when each position of the pattern is used with equal frequency. Here, a binary version of the MNIST dataset [16] is analyzed. Each gray-scale pixel is transformed into a binary value using a threshold value of 0.5. Each image is 28 pixels wide and 28 pixels high, totaling 784 pixels per image. (top-left) An example of a binary MNIST pattern. (top-right) The pixel usage frequency (encoded with gray-scale) of each pixel. (bottom) The same values depicted in the top-right picture, visualized in a single horizontal axis. Notice how the patterns are not sparse ($M = 0.13 \times 784 \approx 100$ when the optimal value is $M^* = \log_2\left(\frac{784}{4}\right) \approx 8$). Furthermore, notice how the patterns are not distributed, as there are some clear predominant areas. For a deeper analysis of the storage of binary MNIST patterns for different threshold values, see the naive baseline in [32].

- The memory can store large amounts of SDRs. Even when the retrieval becomes imperfect as the memory fills up, the noisy retrieved patterns keep the relevant information about the pattern.
- The memory can effectively retrieve noisy cues where active bits are randomly deleted.

Here we complete the analysis by including a decoder module in the pipeline [35] (Fig. 3). This way we can get a visual representation of the SDRs in the several steps of the storage and retrieval process, and evaluate the memory with measures such as the Mean Squared Error (MSE) between the original patterns $P$, and reconstructions $\widehat{P}$: $\text{MSE} = \frac{1}{n}\sum_{i=1}^{n}\left(\frac{1}{784}\sum_{j=1}^{784}\left(P_j^{(i)} - \widehat{P}_j^{(i)}\right)^2\right)$, where 784 is the number of pixels in each image, and $n$ is the number of patterns.

## 2.1. Retrieval

Let us start with the simplest of cases. Here, we present the memory with cues that are identical to those that the memory stored in its learning stage. Recall that since the memory is implementing an auto-associative mapping ($x \rightarrow x$), the goal of the memory is to output a copy of the cue.

The Willshaw Network does not retrieve its stored patterns perfectly. In fact, the average number of bits in the retrieved vector increases by a factor of 4 (Fig. 4(a)) as we store the entire MNIST dataset in the memory. Despite this phenomenon, the MSE between the original images and the reconstructions only increases by a factor of 1.2 (Fig. 4(b)). This result shows that the noise introduced by the memory is non-random and that the retrieved vectors keep the information relevant to the pattern (as originally proposed in [32]). This can be confirmed by visually inspecting the examples of Fig. 5.

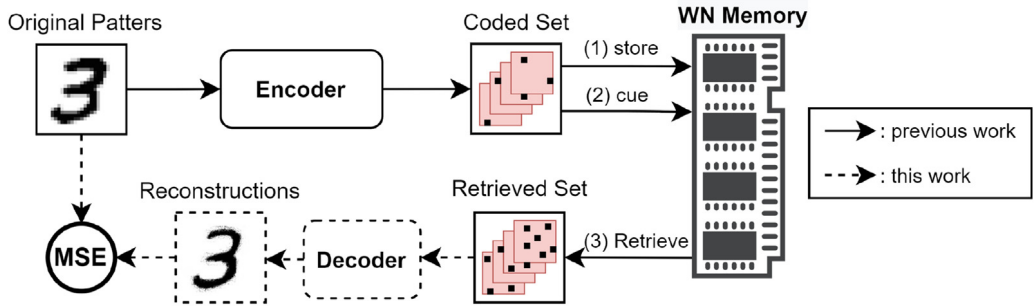

**Fig. 3.** Encoder Decoder and Memory pipeline. First, we encode [32] the visual patterns of the MNIST dataset into SDRs and store them in a Willshaw Network in auto-association. We then cue the trained memory (with either the original WW codes or noisy versions of it) to obtain the retrieved codes. Finally, the WW decoder can be used to obtain a visual representation of the memory's content.

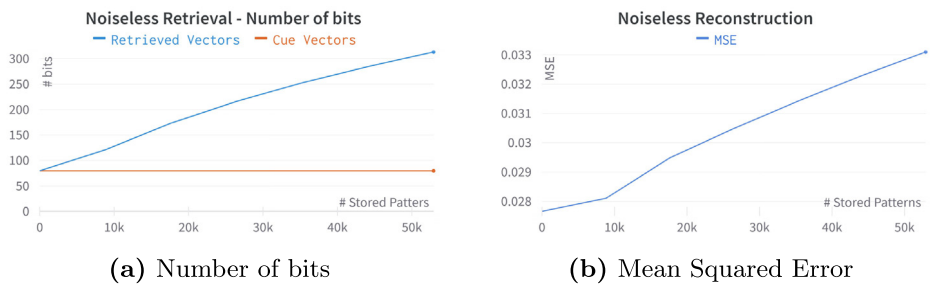

(a) Number of bits

(b) Mean Squared Error

**Fig. 4.** Noiseless Retrieval. Here we analyze the retrieval process of the best-performing sparse codes of [32]. We measure the number of active bits in the retrieval cue, and retrieved vector (a); and the Mean Squared Error between the original MNIST patterns and the reconstructions of the retrieved vectors (b). Notice how, despite the heavy increase in the information in the retrieved vector (a), the reconstruction error does not increase significantly (b).

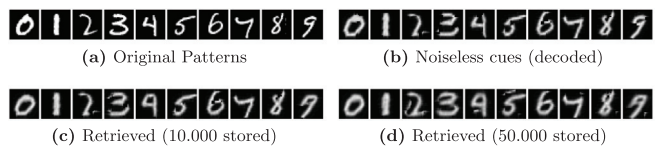

(a) Original Patterns      (b) Noiseless cues (decoded)

(c) Retrieved (10.000 stored)      (d) Retrieved (50.000 stored)

**Fig. 5.** Noiseless retrieval. Notice how the decodings in (b) are very similar to the original patterns (a), with only a few pixels missing due to the lossiness of the encoding process. As the memory feels up (c) the memory adds noise to cues. Most of the noise is around the active pixels of the digit, but some noisy spots start to appear when the memory becomes fuller (d).

## 2.2. Completion

Let us now consider a more practical scenario. Consider that the retrieval cues are incomplete versions of the patterns that are stored in the memory. Formally, the cues $\tilde{X}$ are obtained from the stored patterns $X$ by stochastically deleting each active bit of $X$ with probability $P_{del}$. The goal is to test the content-addressable capability of the model. If the memory has learned a robust auto-associative mapping, it will be able to match the subset of information in the cue with the corresponding stored pattern.

To have a more detailed analysis, we subdivide the MSE into its negative and positive parts: the Mean Squared Error (MSE) due to lost information (when $P_j^{(i)} > \hat{P}_j^{(i)}$), and the MSE due to extra information (when $P_j^{(i)} < \hat{P}_j^{(i)}$) (Fig. 6).

Our results show how the memory can very effectively complete the missing information from the noisy cues. Notice how, the MSE due to loss information sharply drops once we used the memory (compare $x = 0$, with $x = 9.000$ of Fig. 6(a), which corresponds to the first stored batch). The MSE due to extra information increases as the memory is loaded, which is the natural tendency of the retrieval process (notice the same trend in Fig. 4(b)). Fig. 7 illustrates the information completion capabilities of the model.

## 3. Multiple-Modality Architecture

The Willshaw Network (WN) makes no assumption about the data that it stores. Each bit of a pattern that is stored in a WN is interpreted as an informative feature. The position of a particular feature within the pattern, and its meaning/interpretation are irrelevant to the memory. As a result, we can fill the memory with heterogeneous patterns; i.e., patterns containing multiple modalities/types of information, as seen in Fig. 8. In such cases, we are building a Multiple-Modality Willshaw Network (MMWN).

As we have seen before, one of the main strengths of the WN is the ability to complete missing information. A MMWN benefits immensely from this property: During training, the memory will learn how all the different features, from all modalities, are correlated. On retrieval, all the different modalities contribute to the memory's response. If the information from one of the modalities is missing in the retrieval cue, the memory can complete it using the remaining modalities.

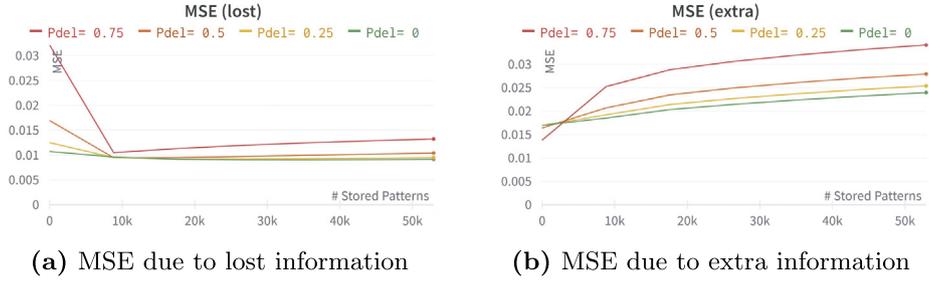

**(a)** MSE due to lost information

**(b)** MSE due to extra information

**Fig. 6.** WW Decoder Reconstruction Error - Noisy (type zero). Here we plot the MSE between the original patterns and the reconstructions. When $x = 0$ we are measuring the MSE of the cue reconstructions. On the rest, we measure the quality of the **reconstructions** that utilize the memory. The green lines correspond to a run without noise, the remaining plots correspond to runs where the retrieval cues are altered by adding **zeros** to the code.

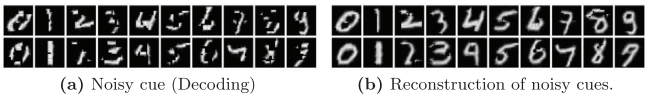

**(a)** Noisy cue (Decoding)      **(b)** Reconstruction of noisy cues.

**Fig. 7.** Completion - examples. The reconstruction of the noisy cues. (b): The reconstruction of the outputs of a memory that stores 10.000 patterns. Notice the effect of the noise ($P_{del} = 0.75$) on the cues: the digits are missing a lot of pixels (a). Using the memory's retrieval to complete the missing information leads to an improvement in the reconstructions (b). Notice also some of the spurious retrievals that occur due to the uncertainty introduced by the noise (bottom-most 4 converges to a 9).

Let us illustrate the strength of this idea with a simple example:

### 3.1. Example of multi-modality retrieval

Consider two patterns ($x^1$ and $x^2$), each represented in two different modalities ($a$ and $b$):

- $x_a^1 = (0, 1)$, $x_b^1 = (0, 0, 1, 1)$
- $x_a^2 = (1, 0)$, $x_b^2 = (1, 1, 0, 0)$

Using the Willshaw training rule (Eq. (1)), we could train two distinct WNs (one for each modality) to store the two distinct modalities of the patterns in auto-association, yielding:

Alternatively, we can concatenate the two modalities of each pattern,

- $x_{ab}^1 = (x_a^1 | x_B^1) = (0, 1 | 0, 0, 1, 1) = (0, 1, 0, 0, 1, 1)$

Classification and Generation of real-world data with an AM Model

$$W_a = \begin{bmatrix} 1 & 0 \\ 0 & 1 \end{bmatrix}, W_b = \begin{bmatrix} 1 & 1 & 0 & 0 \\ 1 & 1 & 0 & 0 \\ 0 & 0 & 1 & 1 \\ 0 & 0 & 1 & 1 \end{bmatrix}$$

- $x_{ab}^2 = (x_a^2 | x_b^2) = (1, 0 | 1, 1, 0, 0) = (1, 0, 1, 1, 0, 0)$

and store the concatenated patterns in auto-association, yielding a single multi-modality matrix:

$$W_{ab} = \begin{bmatrix} 1 & 0 & 1 & 1 & 0 & 0 \\ 0 & 1 & 0 & 0 & 1 & 1 \\ 1 & 0 & 1 & 1 & 0 & 0 \\ 1 & 0 & 1 & 1 & 0 & 0 \\ 0 & 1 & 0 & 0 & 1 & 1 \\ 0 & 1 & 0 & 0 & 1 & 1 \end{bmatrix}$$

Notice two things about the resulting matrix:

- The individual matrices from modalities $a$ and $b$ (in blue and red, respectively) are displayed along the diagonal of the bigger matrix. Each of these matrices represents the **intra-modality correlations**.
- The rest of the matrix (green) has non-zero entries, which represent the **inter-modality correlations**.

The single multi-modal memory has all the capabilities of several single-modal ones. But the additional **inter-modality correlations** provide the memory with extra flexibility.

For instance, if we delete one of the modalities from each pattern to form the following retrieval cues:

- $\bar{x}_{ab}^1 = (x_a^1 | 0) = (0, 1, 0, 0, 0, 0)$
- $\bar{x}_{ab}^2 = (0 | x_b^2) = (0, 0, 1, 1, 0, 0)$

And then compute the memory's response using the two-step retrieval rule (Eq. (2,3)), we get:

- $y_{ab}^1 = (0, 1, 0, 0, 1, 1) = x_{ab}^1$
- $y_{ab}^2 = (1, 0, 1, 1, 0, 0) = x_{ab}^2$

The memory can fully retrieve the missing modality in both cases, which would not be possible if we used $W_a$ and $W_b$ separately.

Please note that the Multiple-Modality framework is just an abstraction: the model is simply storing and retrieving larger binary feature vectors using its basic rules, without any notion of modality. As a consequence, the model is not restricted to the use case of this example, where one modality is completely deleted, and the other used to retrieve it. One could, for instance, provide partial information about both modalities, and use the memory to complete both simultaneously.

Many possible use cases leverage the mechanism of completion. In fact, typical learning tasks (e.g., prediction, classification, and generation) can be seamlessly mapped to a completion problem. For this reason, the proposed model should be seen as a framework and not a tool for specific applications.

In the following section, we will show two examples of typical learning tasks which can be solved under the Multi-Modal framework.

## 4. Experimental analysis

To demonstrate that the MMWN can work in practice, let us use the MNIST dataset to train a MMWN with two modalities and use it for practical applications. Each pattern in the MNIST dataset has two attributes: the label represented by an integer; and a picture of the handwritten digit represented as an array of pixels. Our pre-vious work [32] has shown that a normal WN can efficiently store and retrieve the images of the MNIST dataset. Here, we will use a MMWN that stores both modalities; i.e., both the images and the labels of the MNIST dataset. If the MMWN manages to store both modalities efficiently, we can leverage the information-completion properties of the model for practical applications such as the generation of new patterns and classification.

### 4.1. A simple SDR for integers

To use the label of the MNIST digit (which is an integer between 0 and 9) as a modality, we must first transform it into a binary code. Furthermore, this binary code should be suitable for a WN; i.e., be an SDR.

#### 4.1.1. Noisy X-Hot Encoder

Our proposed encoding strategy is the Noisy X-Hot (NXH) encoding, which is a stochastic version of the X-Hot encoding strategy (a more general version of the well-known 1-Hot codes). The main idea is that $X$ bits in the code will randomly activate with high probability, while the remaining bits will be active with low probability. The result is an X-Hot encoding with some added noise/randomness, hence the name.

Formally, for a label $l$ in the range $\{0, \ldots, L-1\}$, its Noisy-X-hot encoding $e(l)$, with $X$ bits per class, and probabilities $P_{class}$, and $P_{rest}$ (with $P_{class} \gg P_{rest}$), is a binary array of size $L \times X$ given by:

$$e_i(l) = \begin{cases} 1 & \text{with } P_{class} \text{ probability, if } i \in \{z \in \mathbb{Z} | lX \leqslant z < (l+1)X\} \\ 1 & \text{with } P_{rest} \text{ probability, if } i \notin \{z \in \mathbb{Z} | lX \leqslant z < (l+1)X\} \\ 0 & \text{otherwise} \end{cases}$$

(4)

The resulting code can be thought of as a collection of $L$ populations with $X$ neurons, one for each of the $L$ integer values we are encoding. For a given label $l$, its NXH code will have most of its activity on the $l^{th}$ interval. Furthermore, two distinct patterns with identical labels ($l_1 = l_2$) will likely have different NXH encodings ($e(l_1) \neq e(l_2)$) due to the stochastic nature of the NXH. In this way, the NXH code is not only an encoding of the labels but also a **description** which allows us to differentiate between different patterns with the same label.

#### 4.1.2. Noisy X-Hot Decoder

For practical purposes, we must have a mechanism that maps the NXH codes back into the integer value of the label. We can achieve this, by looking at the total activity of all the $L$ intervals of the NXH code and picking the interval with the most activity.

Formally, a NXH code $e$, can be mapped back into its corresponding label $l$ with:

$$l = \arg\max_i \left( \sum_{n=iX}^{(i+1)X} e_n \right)$$

(5)

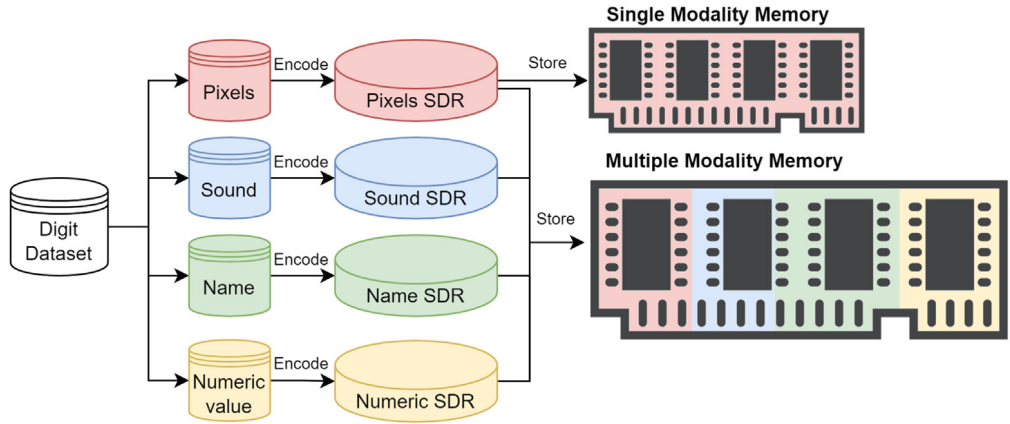

**Fig. 8.** Multiple vs Single modality memory. In this example, we are storing digits in both memories. If we consider a single type of information about the digit, then we require a Single Modality AM (top). If we consider multiple types of information about each pattern, we are using a Multiple Modality AM (bottom).

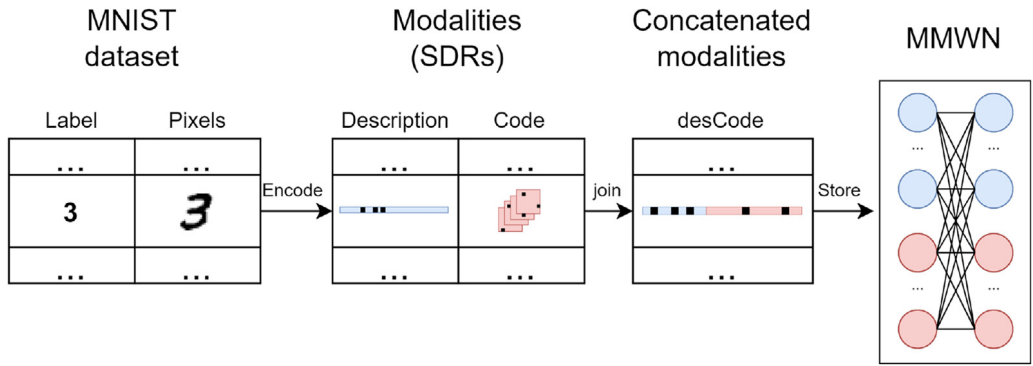

**Fig. 9.** Training Step in a MMWN.

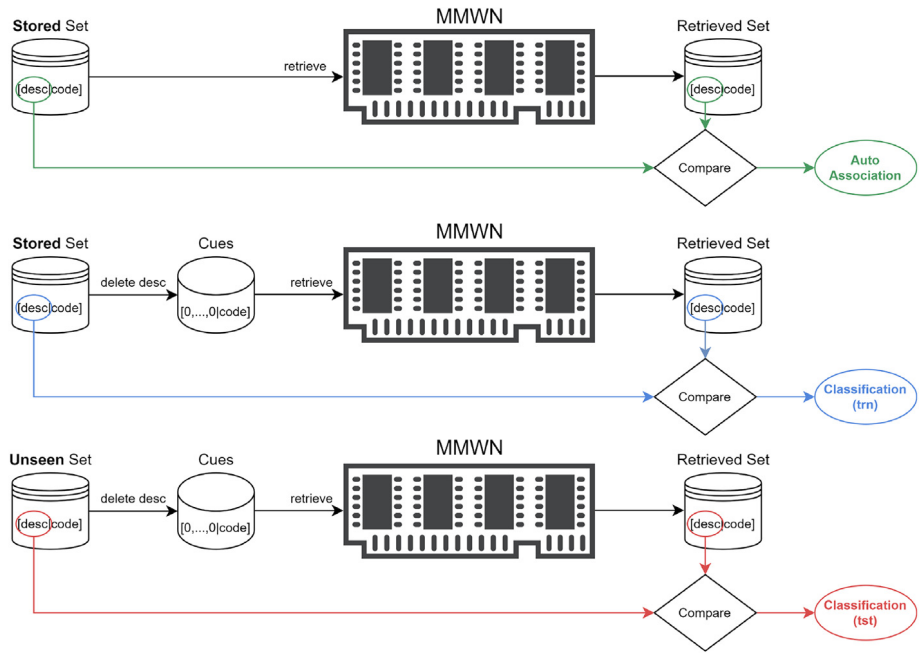

**Fig. 10.** MMWN Classification methodology. To evaluate the MMWN's ability to perform classification we define three distinct tasks of increasing difficulty: (Top): Auto-Association, (Middle): Classification of stored patterns, and (Bottom): Classification of unseen patterns. In all cases, we compare the original description of the pattern (before memory) with the description in the retrieved vector (after memory). For all three tasks, the accuracy score is determined by comparing the integer value of the label before and after the memory which is achieved by decoding the description with Eq. (5).

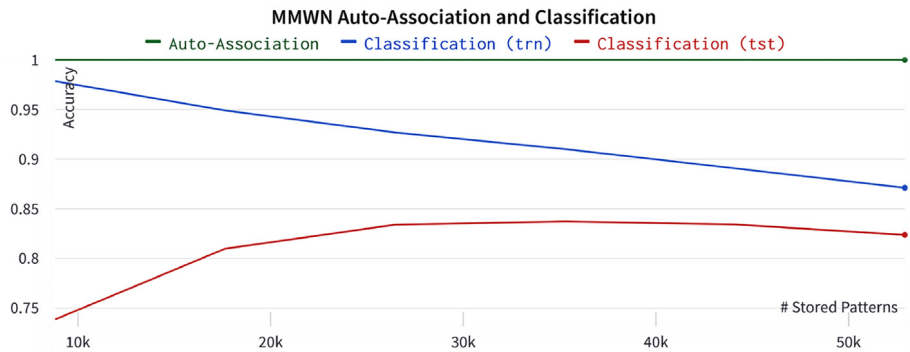

**Fig. 11.** Auto-Association and Classification Results with the MMWN. Here we follow the methodology of Fig. 10 to report the scores of the three tasks as the memory is filled with patterns. For the WW encoder [32], we used the parameters of the best performer in [32]. For the NXH encoder we used: $X = 500$, $P_{class} = 0.5$, and $P_{rest} = 0.0$. (Green): Auto-Association score. Here we are simply measuring the memory's ability to auto-associate the description of DCs. The accuracy is perfect even when the memory is full, indicating that the descriptions are tolerant to the noise that the memory introduces. (Blue): Classification score on train (trn) data (i.e., stored patterns). This score is high but decreases monotonically as the memory is filled up, which is expected since the memory gradually loses the ability to perfectly retrieve the patterns that it stores. (Red): The classification accuracy on test (tst) data (i.e., unseen patterns). This score improves as the memory stores more information because the memory naturally becomes better at generalizing as it stores more information (peeking at 84.04%). However, once the memory becomes too full, this score gets worse.

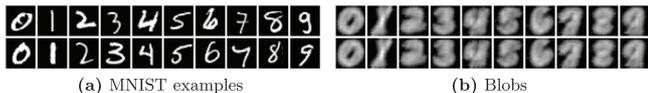

(a) MNIST examples      (b) Blobs

**Fig. 12.** MNIST examples **(a)** and Drawings from memory (blobs) (b). The blobs are obtained when we provide a MMWN with cues that contain the description but are missing the visual modality.

## 4.2. Methodology

There are two key steps when operating an AM: the learning step and the retrieval step. The same is true for a MMWN.

The learning step consists in: (1) encoding the labels and pixels into the descriptions and What-Where codes [32], respectively; (2) concatenating the description and code into what we refer to as a desCode (DC) (short for description and code); and (3) storing the DCs in auto-association, as depicted in Fig. 9.

The Retrieval Step simply consists in showing a cue DC to the trained memory and obtaining a retrieved DC back. As shown in the example of Section 3.1, the MMWN can be used to complete missing modalities using information from another. The flexibility of this framework allows us to manipulate the retrieval cues in different ways to perform different learning tasks.

## 4.3. Classification

If one shows, to a trained MMWN, cues where the description modality has been deleted, the MMWN will complete the missing information, essentially working as a classifier.

To test the MMWN's performance when it comes to classification we defined three tasks of increasing complexity: Auto-association, classification of stored patterns, and classification of unseen patterns (see Fig. 10). Results are reported in Fig. 11.

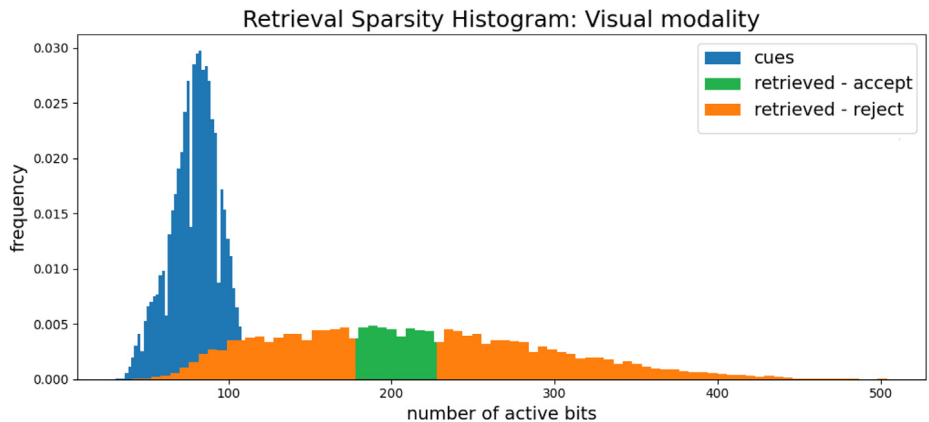

**Fig. 13.** Retrieval Acceptance Interval. Here we used a trained memory to complete noisy versions of stored patterns. The number of bits in the cues, and retrieved vectors is measured and plotted. We can see that both the cues (blue) and retrieved vectors (orange/green) appear to follow a normal distribution. The number of bits in the retrieved vector is indicative of the quality of the retrieval process: Values close to the center of the distribution (green) are a good indicator of a successful retrieval. Values in the tails of the distribution (orange) correspond to poor retrievals, where the cue led to a lack or excess of detected correlation by the memory.

### 4.4. Generation

Inversely to classification, generation with a MMWN is performed by creating a set of cues where the description modality remains intact but the visual modality is deleted. The memory's response will complete the missing information in the visual modality, essentially generating a pattern from a description.

Providing a trained MMWN with a cue where the visual modality has been completely set to zero yields the results in Fig. 12(b). The description modality alone is not able to generate unique patterns that resemble examples from the original dataset, such as those in Fig. 12(a). Instead, "blobs" with no detail, which looks like prototypes of each class, are obtained.

To move away from "blobs" and create more realistic generations, we cannot provide zero information in the visual modality. Instead, we must "seed" the generation process by adding some visual information to the retrieval cue, so that the memory can leverage its information completion capabilities. Furthermore, the visual information provided in the seed should be in accordance with the description that we are generating from (e.g., if we generate from class "zero", we should seed the generation process with some visual features that often occur in patterns of the zero class). The question then becomes how to create these class-dependent generation seeds.

In an ideal scenario, with access to the probability distributions of the visual features for each class, one could sample from these distributions to create artificial cues, and retrieve them in an MMWN to obtain generations. A much more elegant and self-contained alternative is to use the blobs (Fig. 12(b)) as an approximation for the probability distribution of each class: These blobs are very rich in information, containing many more active bits than the average pattern that is stored in memory. Essentially, they are a collection of all the features that are commonly found in examples of the class that originated the blob. If one deletes most of the bits from a blob (which we will denote as "sparsification" throughout the next subsections), the result is a small set of visual features that can be used as a retrieval cue for generation.

### 4.5. Proposed Generation Algorithm

To use a Willshaw Memory as a generative model, an iterative generation algorithm is proposed (see Algorithm 1). The general idea is to iteratively retrieve and sparsify the inputs and outputs of the retrieval process, respectively [35]. In the beginning, the visual modality of the generation cue is empty, but with each iteration, we improve the generation seed and provide the memory with more information. This way, the memory will gradually get closer to a pattern similar to those that it stores. The process stops once the number of bits in the generated code falls within a predefined acceptance interval (see Fig. 13).

An overview of the architecture of the iterative generation method can be seen in Fig. 14(a). Additionally, an illustrative example is provided in Fig. 14(b), 14(c). The first shows how the visual patterns evolve throughout the generation process, while the second monitors the number of active bits. Finally, the results of the iterative generation method can be seen in Fig. 15.

---

**Algorithm 1 Generation with the Willshaw Associative Memory.** The proposed generation process requires a trained Multi-Modal Associative Memory (AM), the class of the pattern we intend to generate ($label$), an acceptance interval ($I_m, I_M$) (see Fig. 13), and two additional parameters ($S_0, S_{inc}$) which control how much information is deleted in the sparsification step of each iteration. Firstly, an empty visual code is concatenated with an encoding of $label$. Then, a cyclical process begins where we retrieve using the memory (retrieval step), and then randomly delete active bits from the memory's response until a level of sparsity $S$ is achieved (sparsification step). The process ends once the sparsity of the output of the retrieval step falls within the acceptance interval $[I_m, I_M]$. With each iteration, we increase $S$ by $S_{inc}$ (i.e., we delete less information with each iteration). As a result, the memory will gradually move from a blob-like pattern (Fig. 12(b)), to a more realistic one (see Fig. 14(b)). The values of $I_m, I_M, S_0$, and $S_{inc}$ are of extreme importance: Since the Willshaw memory uses a one-step retrieval rule, these parameters are what ensures that the algorithm converges towards a solution.

---

**Require:** AM ▷ Trained Multi-Modal Memory
**Require:** $label$ ▷ Label of the class to generate
**Require:** $I_m, I_M$ ▷ Acceptance interval minimum, and Maximum, respectively
**Require:** $S_0, S_{inc}$ ▷ Initial Sparsity and sparsity increment, respectively

$code \leftarrow (0,...,0)$ ▷ Empty image modality
$desc \leftarrow encode(label)$ ▷ See Section 4.1.1 for encoding strategy

$S \leftarrow S_0$
**while** $true$ **do**
  $(des_{gen}|code_{gen}) \leftarrow$ AM.$retrieve((desc|code))$ ▷ Retrieval Step
  **if** $(I_m < activity(code_{gen}) < I_M)$ **then** ▷ Evaluate sparsity
    **break** ▷ Success
  **else**
    $code \leftarrow sparsify(code_{gen}, S)$ ▷ Sparsification step
    $S \leftarrow S + S_{inc}$ ▷ Adjust sparsity for next iteration
  **end if**
**end while**
**return** $decode(code_{gen})$ ▷ Use the decoder to obtain the generation

---

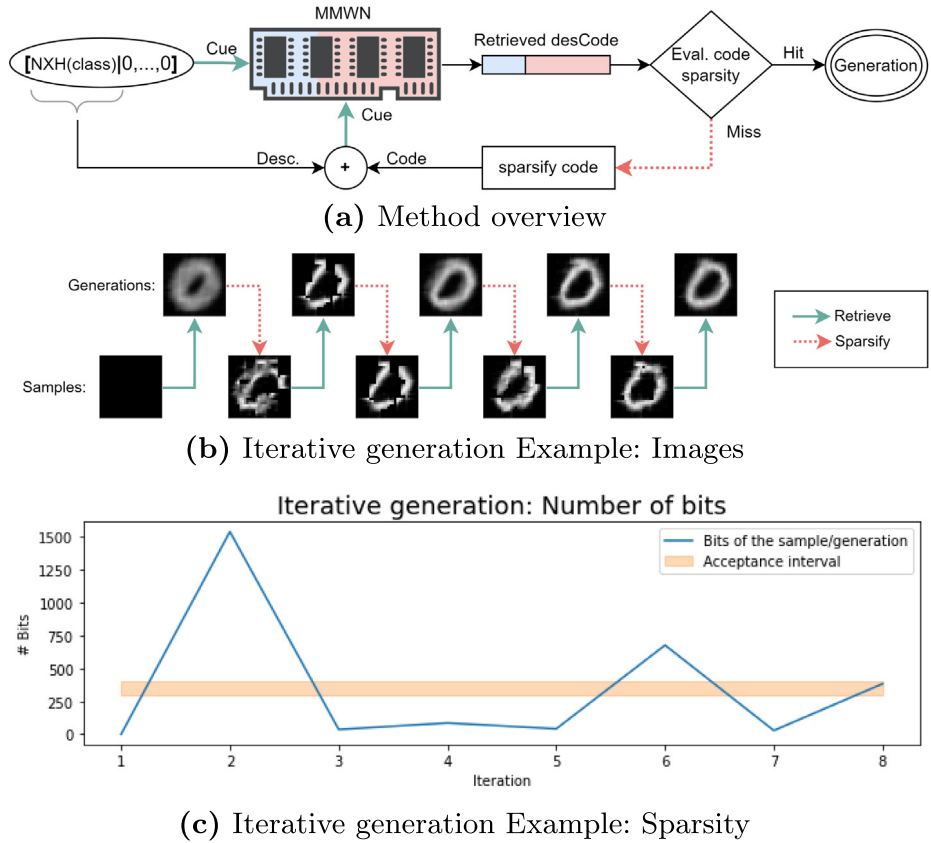

**(a)** Method overview

**(b)** Iterative generation Example: Images

**(c)** Iterative generation Example: Sparsity

**Fig. 14.** Iterative Generation. (a): A visual representation of Algorithm 1. (b): The patterns on the bottom row are retrieval cues that act as "samples" in the generation process. The memory retrieves (green arrows) these samples and produces generations which are displayed on the top row. If the generation's sparsity is outside the acceptance interval, we randomly delete bits from the generated code (dashed red arrows) to obtain a new sample. (c): Here we plot the number of bits throughout the iterative generation process. This graph corresponds to the same example as Fig. 14(b). The first sample has zero active bits since it corresponds to an empty code. The sparsity of the encoding process will oscillate (blue line): When we "sparsify" we bring the number of bits down. On retrieval, the memory will complete the missing information and add more bits. When the output of the memory falls within the acceptance interval (orange) the generation process ends.

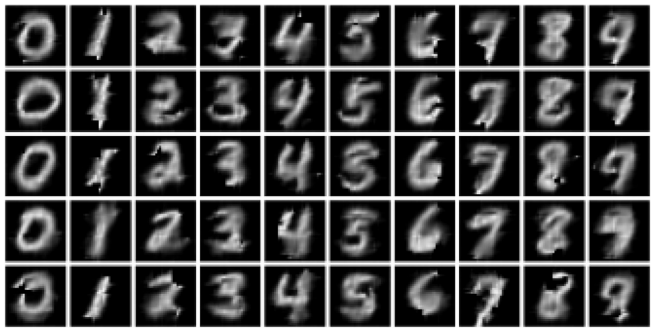

**Fig. 15.** Iterative generation examples. Here we follow the iterative method described in Fig. 14(a)(a) to generate 5 examples from each MNIST class with the MMWN. All generations look like realistic examples from the class that they belong to. Furthermore, there is some variance between generations of the same class. These results were obtained with an acceptance interval of [400, 500] bits.

## 5. Conclusion

In this paper, we use a Multiple-Modality framework for the basic Willshaw model and demonstrate how it can be applied to real-world data. The resulting system is a computational model of brain functions which learns by storing several modalities (e.g., visual, textual, etc.) of each pattern simultaneously. After storage, the model's completion capabilities can be leveraged to infer missing modalities when just a subset is perceived, thus serving a framework for learning tasks. To demonstrate the idea, we evaluated the model on the MNIST dataset of handwritten digits. Using sparse encoding prescriptions for the image [33] and the numeric value (Section 4.1.1) of the patterns, we were able to perform several learning tasks which are crucial in intelligent systems, namely Retrieval, Completion, Classification, and Generation. Additionally, we propose an iterative algorithm (Algorithm 1) that greatly improves the generation capabilities of the model. All the reported results follow a simple encoder-memory-decoder architecture, where the memory is the simple Willshaw Associative Memory, which trains in a single pass through the data, using a simple Hebbian rule.

Despite being evaluated on traditional Machine Learning tasks, the model's performance on individual tasks is not expected to outperform state-of-the-art deep architectures. After all, the proposed model is intended to be a general information processing framework, not designed for specific applications.

This work is a first step, where an idea is proposed and its feasibility demonstrated. Therefore, the results reported here could certainly be improved with future research efforts, e.g., by applying the iterative retrieval method (Algorithm 1) to tasks other than generation. Additional research possibilities include performing other completion-based learning tasks (e.g., regression, time-series prediction, anomaly detection, etc.) under the Multiple-Modality framework; storing several modality-sharing-datasets simultaneously, which is possible since the system is not domain-bound; and scaling up the model to include additional modalities, which entails building sparse encoding prescriptions as well.

While far from a replica of a biological system, the design of our solution is heavily constrained by biology. Therefore, its successful application in artificial intelligent tasks is exciting, as it might shed some light on the mechanisms that operate in biological memories.

## CRediT authorship contribution statement

**Rodrigo Simas:** Methodology, Software, Investigation, Writing - original draft, Writing - review & editing, Visualization. **Luis Sa-Couto:** Methodology, Software, Supervision, Conceptualization. **Andreas Wichert:** Supervision, Conceptualization.

## Data availability

Data will be made available on request.

## Declaration of Competing Interest

The authors declare that they have no known competing financial interests or personal relationships that could have appeared to influence the work reported in this paper.

## Acknowledgments

This work was supported by national funds through Fundação para a Ciência e Tecnologia (FCT) with reference UIDB/50021/2020 and through doctoral grant SFRH/BD/144560/2019 awarded to the second author. The funders had no role in study design, data collection and analysis, decision to publish, or preparation of the manuscript.

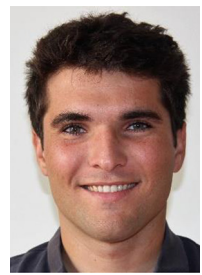

**Rodrigo Simas** is a recent MSc graduate in Computer Science - Instituto Superior Técnico, University of Lisbon. He did his research for his Dissertation project under the Supervision of Prof. Andreas Whichert and Co-supervision of Luis Sá Couto.

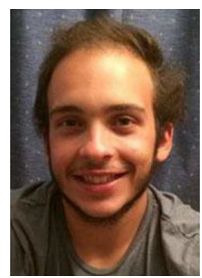

**Luis Sá Couto** studied computer science at the Department of Computer Science and Engineering, University of Lisbon, where he graduated in 2018. Since then he is a PhD student under the supervision of Prof Andreas Wichert with the topic Employing brain inspired principles towards better and more general learning of simple vision. He lectures practical classes in AI and Machine Learning

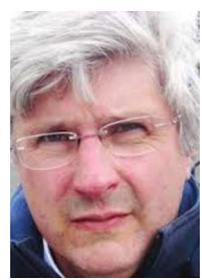

**Andreas Wichert** studied computer science at the University of Saarland, where he graduated in 1993. Afterwards, he became a PhD student at the Department of Neural Information Processing, University of Ulm. Since 2006 he is Assistant Professor at Department of Computer Science and Engineering, University of Lisbon where he is as well lecturing about machine learning and quantum computation. His research focuses on neuronal networks, cognitive systems and quantum computation.